\documentclass[runningheads]{llncs}

\usepackage[mobile]{eccv}

\usepackage{eccvabbrv}

\usepackage{graphicx}
\usepackage{booktabs}
\usepackage{wrapfig}

\usepackage[accsupp]{axessibility}  

\usepackage{hyperref}

\usepackage{orcidlink}

\usepackage{comment}

\usepackage{multirow}
\usepackage{array}
\usepackage{acronym}
\usepackage{tikz}
\usepackage[ruled,vlined]{algorithm2e}

\definecolor{CellBck}{gray}{0.92}
\definecolor{CellBlue}{RGB}{240, 240, 255}
\definecolor{OracleColor}{gray}{0.5}
\definecolor{AugColor}{RGB}{25, 175, 25}
\definecolor{DecrColor}{RGB}{175, 25, 25}

\definecolor{OracleColor}{RGB}{100, 100, 100}
\definecolor{Methodcolor}{gray}{0.1}

\definecolor{AlgCommentColor}{RGB}{31, 153, 46}

\newcommand{\cbg}{\cellcolor{CellBck}}
\newcommand{\cbgb}{\cellcolor{CellBlue}}
\newcommand{\incr}[1]{{\tiny \color{AugColor} $\uparrow$ #1}}
\newcommand{\decr}[1]{{\tiny \color{DecrColor} $\downarrow$ #1}}

\newcommand{\oracle}{{{\color{Methodcolor} Oracle}}}
\newcommand{\baseline}{{{\color{Methodcolor} Source-Only}}}
\newcommand{\join}{{{\color{Methodcolor} Joint-Train}}}
\newcommand{\tent}{{{\color{Methodcolor} TENT}}}
\newcommand{\dua}{{{\color{Methodcolor} DUA}}}
\newcommand{\xmuda}{{{\color{Methodcolor} xMUDA}}}

\newcommand{\mmtta}{{{\color{Methodcolor} MM-TTA}}}

\newcommand{\tttcont}{{{\color{Methodcolor} AdaContrast}}}
\newcommand{\ttkdo}{{{\color{Methodcolor} TTT-KD-O}}}
\newcommand{\ttkd}{{{\color{Methodcolor} TTT-KD}}}

\newcommand{\scannet}{\textsc{ScanNet} }
\newcommand{\sdis}{\textsc{S3DIS} }
\newcommand{\matterportd}{\textsc{Matterport3D} }
\newcommand{\mattershort}{\textsc{Matt3D}}
\newcommand{\atwodtwo}{\textsc{A2D2} }
\newcommand{\kitti}{\textsc{Kitti} }
\newcommand{\skitti}{\textsc{SemanticKitti} }

\acrodef{KD}[KD]{Knowledge Distillation}
\acrodef{LLM}[LLM]{Large Language Models}
\acrodef{UDA}[UDA]{Unsupervised Domain Adaptation}
\acrodef{TTA}[TTA]{Test-Time Adaptation}
\acrodef{TTT}[TTT]{Test-Time Training}
\acrodef{MAE}[MAE]{Masked Auto Encoding}
\acrodef{TTKD}[TTKD]{Test-Time Knowledge Distillation}
\acrodef{SGD}[SGD]{Stochastic Gradient Descent}
\acrodef{mIoU}[mIoU]{mean Intersection over Union}
\acrodef{mAcc}[mAcc]{mean per-class Accuracy}
\acrodef{ODD}[OOD]{out-of-distribution}
\acrodef{ID}[ID]{in-distribution}
\acrodef{MLP}[MLP]{Multi-layer Perceptron}

\begin{document}

\title{TTT-KD:  Test-Time Training\\for 3D Semantic Segmentation through\\Knowledge Distillation from Foundation Models} 

\titlerunning{TTT-KD}

\author{
\textbf{Lisa Weijler\inst{1}}\quad
\textbf{M. Jehanzeb Mirza\inst{2,3}}\quad
\textbf{Leon Sick\inst{4}}\quad
\textbf{Can Ekkazan\inst{5}}\quad
\textbf{Pedro Hermosilla\inst{1}}\\
{\institute{$^1$TU Wien, Austria. \quad $^2$ICG, TU Graz, Austria. \quad $^3$CDL-EML. \quad $^4$Ulm University \quad $^5$Yildiz Technical University}}}

\authorrunning{L.~Weijler et al.}

\maketitle



\begin{abstract}
    \ac{TTT} proposes to adapt a pre-trained network to changing data distributions on-the-fly. 
    In this work, we propose the first \ac{TTT} method for 3D semantic segmentation, \textbf{TTT-KD}, which models \ac{KD} from foundation models (\eg DINOv2) as a self-supervised objective for adaptation to distribution shifts at test-time. 
    Given access to paired image-pointcloud (2D-3D) data, we first optimize a 3D segmentation backbone for the main task of semantic segmentation using the pointclouds and the task of 2D~$\to$~3D \ac{KD} by using an off-the-shelf 2D pre-trained foundation model. 
    At test-time, our TTT-KD updates the 3D segmentation backbone for each test sample, by using the self-supervised task of knowledge distillation, before performing the final prediction.
    Extensive evaluations on multiple indoor and outdoor 3D segmentation benchmarks show the utility of TTT-KD, as it improves performance for both \ac{ID} and \ac{ODD} test datasets.
    We achieve a gain of up to 13\,\% mIoU (7\,\% on average) when the train and test distributions are similar and up to 45\,\% (20\,\% on average) when adapting to \ac{ODD} test samples. 
\end{abstract}    

\begin{figure}
    \centering
    \includegraphics[width=\textwidth]{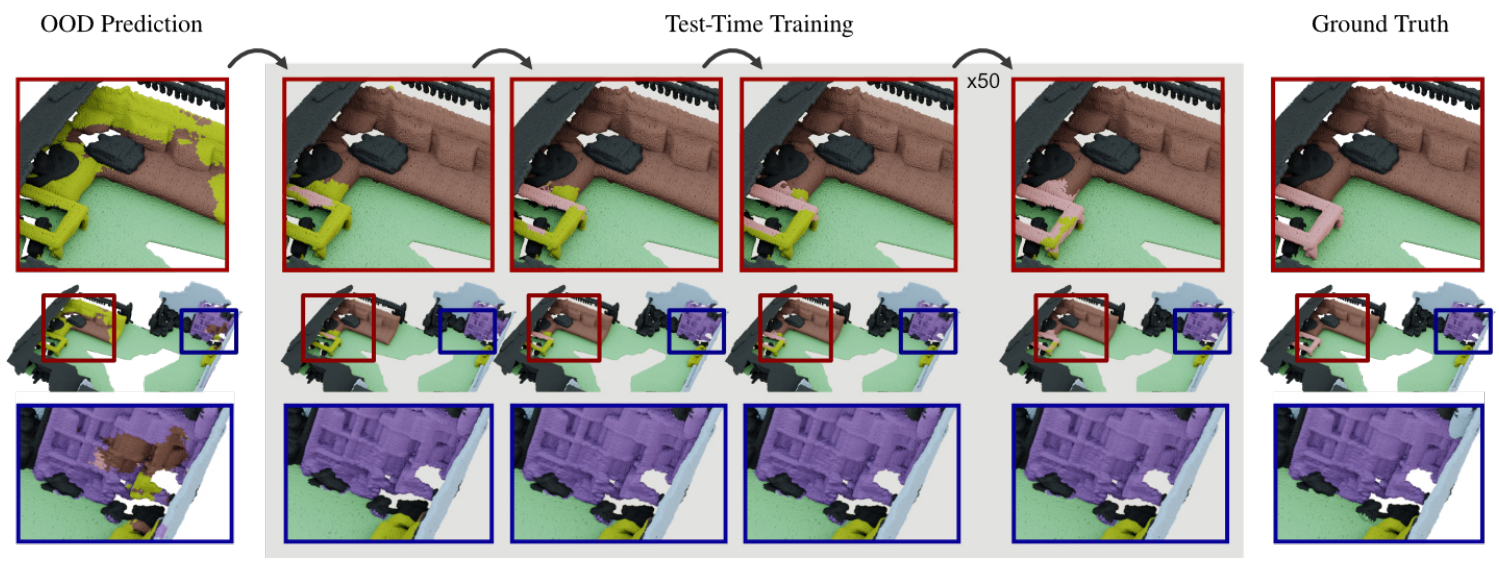}
    \captionof{figure}{\textbf{TTT-KD.} We propose the first test-time training method for 3D semantic segmentation which adapts to distribution shifts at test time. As shown in the illustration above, our method is able to adapt to Out-of-Distribution (OOD) scenes (\scannet) where the model was not trained on (\sdis). Iteratively, via knowledge distillation from 2D foundation models, our algorithm adjusts the weights of the network, progressively improving prediction with each step \textbf{(top)}. Moreover, our approach is able to significantly improve the predictions with even a single step while maintaining the quality of those without degradation over multiple steps \textbf{(bottom)}.}
    \label{fig:teaser}
\end{figure}

\section{Introduction}
\label{sec:intro}

3D semantic segmentation represents a fundamental benchmark for neural networks that process pointclouds~\cite{Wang2023octformer, hermosilla2023pne, wu2022pointtrans, choy20194d}.
In this task, the model's primary objective is to predict the semantic label of each point in the scene.
Successful execution of this task requires a profound comprehension of scene objects and their precise spatial localization.
Despite the recent success obtained by different models for the task of 3D semantic segmentation, the generalization of these models on different data sets still remains an open problem.
This generalization gap can be provoked due to a variety of reasons:
sensors used during acquisition, reconstruction algorithms used to obtain the pointcloud, inherent noise on the point coordinates, colors, and normals, or even the different scene compositions.

One way to bridge the domain gap is to label the pointcloud sequences from different datasets and train the network in a supervised manner on this data~\cite{wu2023ppt}.
However, labeling can incur huge monetary costs and manual effort.
To avoid these challenges several works suggested to adapt the network in an unsupervised manner to the \ac{ODD} data.
A popular paradigm is \ac{UDA}, where the network is trained jointly on the labeled source domain and unlabeled target domain, with the goal of learning an invariant feature representation for both domains.
Many works~\cite{saltori2020sfuad,yi2021complete,saltori2022cosmix,saltori2023compositional,shaban2023lidaruda, cao2023mopa}, have proposed \ac{UDA} approaches for 3D semantic segmentation for outdoor pointclouds, but countering domain shifts for indoor scenes is relatively less studied.
Moreover, in real-world scenarios, there could rarely be situations where the target domain is known in advance, rendering these methods unsuitable.

To forego the need for access to the target domain data,~\ac{TTA} algorithms~\cite{shin2022mmtta, liang2020we, wang2022continual} instead propose to adapt the network weights at test-time, more generally with some posthoc regularization.
For effective adaptation, these works usually require constrained optimization of the network parameters, for example, only updating the affine parameters of the normalization layers. 
However, this could be insufficient to adapt to severe domain shifts.
Moreover, these methods also rely on larger batch sizes for adaptation, making their applicability to large indoor 3D scenes challenging.

Sharing the philosophy with \ac{TTA} of adapting the network weights at test-time but differing in its application and methodology, \ac{TTT} proposes to first train a network jointly for the main (downstream) task and a self-supervised auxiliary objective.
At test-time, given a (single) pointcloud sample, \ac{TTT} adapts the network weights \textit{independently} for each pointcloud sample by using the self-supervised objective, and then performs inference with the adapted network weights.
Recently, MATE~\cite{mirza2023mate} proposes to use \ac{MAE}~\cite{pang2022masked} task as a self-supervised objective for adaptation to \ac{ODD} pointclouds at test-time, for the task of pointcloud classification. 
However, its architecture makes it unsuitable for application to dense prediction tasks, such as 3D semantic segmentation, and it was only tested on synthetic domain shifts.

In this work, we propose the first \ac{TTT} algorithm for the task of 3D semantic segmentation, TTT-KD, which models 2D $\rightarrow$ 3D \ac{KD} from foundation models as a self-supervised objective.
During training, our method receives a 3D pointcloud as input and a set of 2D images of the same scene with point-pixel correspondences.
A 3D backbone processes the 3D pointcloud generating a set of 3D per-point features.
These 3D features are then used to predict the semantic label of the points and also to perform 2D $\rightarrow$ 3D \ac{KD} from a 2D foundation model, DINOv2~\cite{oquab2023dinov2} in most of our experiments.
At test-time, given a test pointcloud with its corresponding images, we adapt the network's weights by taking several gradient descent steps on the self-supervised task of \ac{KD}.
Since the 3D backbone has learned a joint feature space for the main segmentation task and the self-supervised \ac{KD} task, improving predictions on the \ac{KD} task leads to large improvements in the semantic segmentation task.
Our algorithm does not make any assumption of the target domain and, therefore, it is able to adapt to it by processing individual scenes at a time.
Our extensive evaluation shows that our algorithm not only leads to large improvements for \ac{ODD} datasets (see \cref{fig:teaser}), up to $17$ in \ac{mIoU}, but also provides significant improvements on in-distribution datasets, up to $8.5$ in \ac{mIoU}.
\section{Related Work}
\label{sec:sota}
Our TTT-KD is related to works which study Unsupervised Domain Adaptation (UDA), Test-Time Adaptation (TTA), and more closely to works which propose methodologies for Test-Time Training (TTT).

\paragraph{Unsupervised Domain Adaptation.}
UDA aims to train a network in order to bridge the gap between the source and target domains while having access to labeled data from the source domain and unlabeled data from the target domain. 
For the task of pointcloud classification, PointDAN~\cite{qin2019pointdan} proposes to learn domain invariant features between the source and target domain with the help of adversarial feature alignment~\cite{dann}. 
Liang~\etal~\cite{liang2022point} propose to learn an invariant feature space for the source and target domain by using self-supervision. 
More specifically, they propose to predict the masked local structures by estimating cardinality, position and normals for the points in the source and target domains, while Shen~\etal~\cite{shen2022domain} propose to use implicit functions coupled with pseudo-labeling for UDA.
For 3D object detection, some approaches~\cite{wang2020train, malic2023sailor} rely on statistical normalization of the anchor boxes in the source and target domain for UDA. 
Lehner and Gasperini~\etal~\cite{lehner20223d} propose to use adversarial augmentations, while pseudo-labeling is employed by~\cite{luo2021unsupervised, fruhwirth2021fast3d, yang2021st3d, saltori2020sfuad}.
The task of 3D semantic segmentation has also been studied extensively in the context of UDA. 
Yi~\etal~\cite{yi2021complete} propose to synthesize canonical domain points, making the sparse pointcloud dense, before performing segmentation.
Saltori~\etal~\cite{saltori2023compositional} propose a pseudo-labeling approach by mixing the source and target domains.
xMUDA~\cite{jaritz2019xmuda} proposes a multi-modal (two-stream) learning approach between 3D and 2D networks, they perform UDA by minimizing the discrepancy between the feature space of the two streams and self-training through pseudo-labeling. 
Some other approaches also rely on a multi-modal setup and achieve UDA by increasing the number of samples used from 2D features by increasing the 3D to 2D correspondences~\cite{peng2021stod}, employing contrastive learning~\cite{xing2023contrastuda} or leveraging SAM~\cite{kirillov2023segany} for obtaining reliable dense 2D annotations~\cite{cao2023mopa}. 
Although UDA offers an efficient solution for adaptation to distribution shifts, still it requires advanced knowledge about the target distribution and requires access to the unlabeled data as well. 
However, in real-world scenarios, it is often the case that such luxury is not affordable. 
Distribution shifts can occur~\emph{on-the-fly} and can be unpredictable.
Thus, a more practicable solution is to adapt the network weights whenever changing data distributions are encountered, which is put forward by TTT. 
\paragraph{Test-Time Adaptation.}
TTA does not alter the training procedure of the network but instead proposes post hoc regularization for adaptation to distribution shifts at test-time.
For the image domain, some approaches rely on statistical correction to adapt the network at test-time, generally by adapting the means and variance estimates (of the Batch Normalization layer~\cite{ioffe2015batch}) to the OOD test data~\cite{lim2023ttn,mirza2022dua}.
TENT~\cite{wang2020tent} proposes to adapt to distribution shifts at test-time by minimizing the Shannon Entropy~\cite{shannon1948mathematical} of predictions and adapts only the scale and shift parameters of the normalization layers in the network. 
The problem of TENT~\cite{wang2020tent} to require larger batch sizes is solved by MEMO~\cite{zhang2022memo}, which augments a single sample multiple times and minimizes the marginal output distribution over the augmented samples. 
Niu~\etal~\cite{niu2023towards} proposes a sharpness aware entropy minimization method for adaptation to distribution shifts in the wild. 
One group of TTA methods also rely on self-training. 
T3A~\cite{iwasawa2021test} casts TTA as a prototype learning problem and replaces a classifier learned on the source dataset with pseudo-prototypes generated on-the-fly for the test batch. 
AdaContrast~\cite{chen2022contrastive} uses contrastive learning with a momentum encoder to adapt to distribution shift on-the-fly.
MM-TTA~\cite{shin2022mmtta} uses 2D-3D multi-modal training for test-time adaptation but only adapts batch normalization affine parameters with a pseudo-labeling strategy.
Liang~\etal~\cite{liang2020we} also relies on pseudo-labeling and entropy minimization of individual predictions but also encourages maximizing the entropy over predicted classes over the entire dataset. 
CoTTA~\cite{wang2022continual} also relies on pseudo-labeling and proposes continual test-time adaptation, where they learn different distribution shifts at test-time in a continual manner. 
Similarly, other continual TTA methods include~\cite{dobler2023robust, eata}. 
Another group of methods also relies on consistency of predictions~\cite{boudiaf2022parameter}, or statistics between the train and test data distributions~\cite{mirza2023actmad,lin2023video}.
We port AdaContrast~\cite{chen2022contrastive}, DUA~\cite{mirza2022dua} and TENT~\cite{wang2020tent} to the task of 3D semantic segmentation and, together with MM-TTA~\cite{shin2022mmtta}, choose them as representative methods for TTA. 
Empirically, our TTT-KD outperforms these methods comprehensively on all the benchmarks we test on. 
\paragraph{Test-Time Training.}
TTT first proposes to train the network jointly for the main downstream task (\eg, 3D Semantic Segmentation in our case) and a self-supervised objective.
At test-time, it adapts the network weights, for a single OOD sample as it is encountered, by using the self-supervised objective -- usually by taking multiple gradient steps for each sample.
TTT methods are usually strictly~\emph{inductive} in nature,~\ie, they adapt the network weights on a single sample only, whereas, TTA methods do not adhere to this restriction.
For the image domain, there are two TTT works that differ \emph{w.r.t} the self-supervised objectives they employ for adaptation. 
The first TTT~\cite{sun2020test} method (which popularized the name) employs rotation prediction~\cite{gidaris2018unsupervised} as its self-supervised task.
Unfortunately, this approach is difficult to adapt to 3D semantic segmentation, since 3D scenes do not have a canonical orientation, and training with random SO(3) rotations usually leads to a degradation in the resulting performance.
The second TTT approach, TTT-MAE~\cite{gandelsman2022test}, uses the task of image reconstruction through masked auto-encoders (MAE)~\cite{he2022masked}. 
MATE~\cite{mirza2023mate} proposes a TTT method, which also employs the MAE objective (PointMAE~\cite{pang2022masked} for pointclouds) for adapting to distribution shifts in pointclouds. 
MATE shows impressive performance for the task of pointcloud classification but uses an architecture that is unsuitable for 3D semantic segmentation, which is the focus of our work. 

In this paper, we propose a TTT method which models (2D $\to$ 3D) Knowledge Distillation from foundation models (\eg, DINOv2~\cite{oquab2023dinov2}) as a self-supervised objective for adaptation to distribution shifts at test-time, for the task of 3D semantic segmentation. 
Similar to other TTT works, our TTT-KD is also strictly inductive in nature and adapts on a single pointcloud sample by performing multiple gradient steps for effective adaptation.

\begin{figure}[t]
    \centering
    \includegraphics*[width=\linewidth]{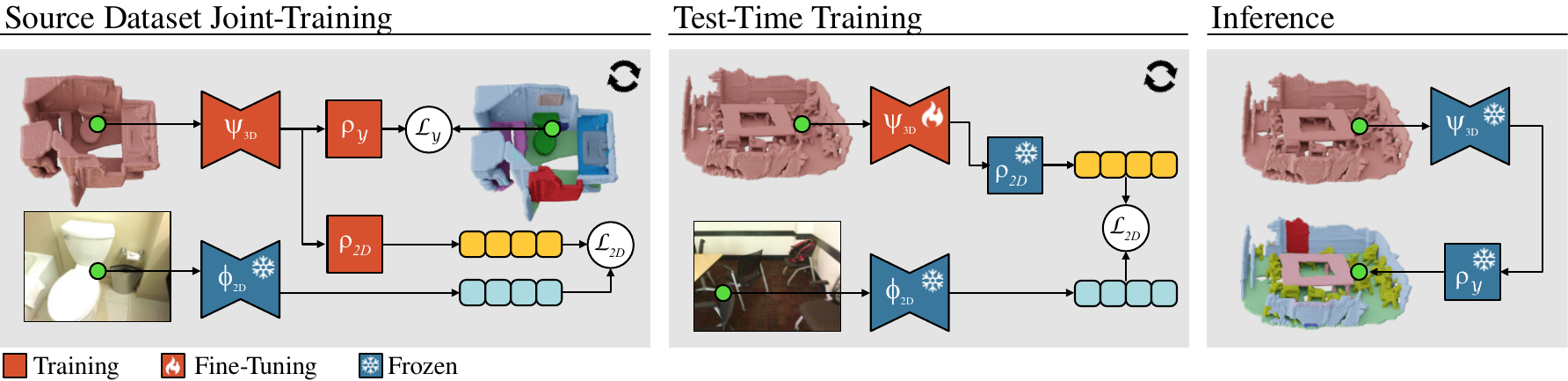}
    \caption{Given paired image-pointcloud data of a 3D scene, \ttkd, during \textbf{joint-training}, optimizes the parameters of a point or voxel-based 3D backbone, $\psi_{3D}$, followed by two projectors, $\rho_{\mathcal{Y}}$ and $\rho_{2D}$.
    While $\rho_{\mathcal{Y}}$ predicts the semantic label of each point, $\rho_{2D}$ is used for knowledge distillation from a frozen 2D foundation model, $\phi_{2D}$.
    During \textbf{test-time training}, for each test scene, we perform several optimization steps on the self-supervised task of knowledge distillation to fine-tune the parameters of the 3D backbone. 
    Lastly, during \textbf{inference}, we freeze all parameters of the model to perform the final prediction.
    By improving on the knowledge distillation task during \ac{TTT}, the model adapts to out-of-distribution 3D scenes different from the source data the model was initially trained on.}
    \label{fig:overview}
\end{figure}

\section{Methods}
\label{sec:methods}

Our algorithm jointly trains a 3D model on the semantic segmentation task and 2D $\rightarrow$ 3D \ac{KD} as a secondary self-supervised task.
In order to be robust to domain shifts, for each scene during testing, we perform a few steps of gradient descent on the \ac{KD} task before we freeze the model to perform the final prediction on the segmentation task.
In this section, we explain the three phases of our method:
\emph{Joint Training}, \emph{Test-Time Training}, and \emph{Inference} (see \cref{fig:overview}).

\subsection{Input}
Our method assumes as input sets of the form $(\mathcal{X}, \mathcal{F}, \mathcal{Y}, \mathcal{I}, \mathcal{U})$, where $\mathcal{X} \in \mathbb{R}^{N \times 3}$ are the spatial coordinates of the $N$ points representing the scene, $\mathcal{F} \in \mathbb{R}^{N \times F}$ are the features associated with each point, $\mathcal{Y} \in \{ 
 0 , 1 \}^{N \times C}$ are per-point semantic labels, $\mathcal{I} \in \mathbb{R}^{I \times W \times H \times 3}$ are a set of $I$ images of the same scene, and $\mathcal{U} \in \mathbb{R}^{I \times N \times 2}$ are the pixel coordinates of each pair of point in $\mathcal{X}$ and image in $\mathcal{I}$.
Note that not all points are projected on all images, and some points of the scene might not be projected on any image.

\subsection{Joint Training}

\paragraph{3D backbone.}
During training, we process each pointcloud $\mathcal{X}$ with a 3D backbone $\psi_{3D}$ to generate semantically relevant 3D features per-point, $F^{3D}$.
Our method is agnostic to the backbone used and works, as we will show later, with voxel-based and point-based architectures.

\paragraph{2D foundation model.}
At the same time, we process all images of the 3D scene, $\mathcal{I}$, with a model $\phi_{2D}$ capable of generating semantically relevant 2D features, $F^{2D}$.
This foundation model is pre-trained in a self-supervised manner on millions of images and remains fixed during the whole training procedure.
As we will show in the ablation studies, our method is also agnostic to the foundation model used and can be used with any off-the-shelf foundation model.

\paragraph{Learning objective.}
Our learning objective is a multi-task objective where, from the 3D features $F^{3D}$, we aim to predict the semantic label of each point, $\hat{\mathcal{Y}}$, and the associated average 2D feature $\hat{F}^{2D}$ over all the images.
Therefore, our algorithm minimizes a combination of two losses:
\begin{align}
    \mathcal{L}_{\mathcal{Y}} &= \mathbb{E}_{x \sim \mathcal{X}} \left[ - \sum_c^C \mathcal{Y}_{x,c} \log (\hat{\mathcal{Y}}_{x,c})\right] \nonumber\\
    \mathcal{L}_{2D} &= \mathbb{E}_{x \sim \mathcal{X}, i \sim \mathcal{I}} \left[ - \frac{\hat{F}^{2D}_x}{\| \hat{F}^{2D}_x\|} \cdot \frac{F^{2D}_i(\mathcal{U}_{x, i})}{\| F^{2D}_i(\mathcal{U}_{x, i})\|} \right] \nonumber
\end{align}
\noindent where $\mathcal{L}_{\mathcal{Y}}$ is the cross-entropy loss between the predicted labels $\hat{\mathcal{Y}}$ and the ground truth labels $\mathcal{Y}$, and $\mathcal{L}_{2D}$ is the knowledge distillation loss defined as the expected cosine similarity between the normalized per point features $\hat{F}^{2D}$ and image features $F^{2D}$, sampled at the pixel position defined by the mapping $\mathcal{U}$.
To estimate $\mathcal{L}_{\mathcal{Y}}$ during training, we compute the average cross-entropy loss of all the points within the batch.
However, since estimating $\mathcal{L}_{2D}$ is more expensive, we randomly sample points $x$ and images $i$ to fill a certain budget per batch.

\paragraph{Feature projection.}
In order to learn a common 3D feature space $F^{3D}$ with these competing objectives without hampering the predictions on the main task, we transform the 3D features to $\hat{\mathcal{Y}}$ and $\hat{F}^{2D}$ with two separate projectors, $\rho_{\mathcal{Y}}$ and $\rho_{2D}$ respectively.
In practice, these projectors are two simple \ac{MLP}.
During training, we optimize the parameters of the 3D backbone, $\psi_{3D}$, and the two projectors, $\rho_{\mathcal{Y}}$ and $\rho_{2D}$, whilst the parameters of the foundation model, $\phi_{2D}$, remain fixed.

\subsection{Test-Time Training}

Contrary to the standard testing phase in other algorithms, in which the parameters of the model are frozen, our algorithm, for each \ac{ODD} scene, slightly modifies the parameters of the model before performing the final prediction.
In particular, we freeze the parameters of the projectors $\rho_{\mathcal{Y}}$ and $\rho_{2D}$, and fine-tune all parameters of $\psi_{3D}$ while fixing the mean and standard deviation of the batch normalization layers.
In particular, we perform several gradient descent steps minimizing the knowledge distillation loss, $\mathcal{L}_{2D}$, for which no labels are required.
Since both projectors have learned to perform predictions from a common feature space, $F_{3D}$, and both projectors aim to predict semantically relevant information, modifying these features to improve $\mathcal{L}_{2D}$ also improves the predictions on the primary segmentation task.
Since we process single scenes, contrary to existing test-time adaptation approaches, we do not update the mean and standard deviation of the batch normalization layers.
Therefore, we are not forced to synthetically increase the batch size with data augmentations, which might be prohibitive for large scenes composed of millions of points.

\subsection{Inference}
Once the test-time training phase has finished, we freeze all parameters of our model and perform the final prediction on the segmentation task.
Following previous works~\cite{sun2020test,mirza2023mate}, we experiment with two variants of our method:

\paragraph{Offline (TTT-KD).}
In this setup, we perform several gradient descent steps for each test scene independently. 
Once the \ac{TTT} phase has finished, we predict the per-point class for the current scene and then we discard the parameter updates before processing the next test scene.

\paragraph{Online (TTT-KD-O).}
In this setup, we only perform one optimization step for each test scene but we keep the parameter updates between consecutive scenes.
Although this approach does not fully adapt to a single scene, it requires less computational resources while, as we will show later, achieving significant improvements over the baselines.

\section{Results}
\label{sec:results}

In this section, we describe the experiments carried out to validate our methods.
In particular, we tested our \ttkd algorithm on two different 3D semantic segmentation setups: indoor and outdoor 3D semantic segmentation.
While indoor 3D semantic segmentation provides an ideal setup for our algorithm, in which each pointcloud is paired with multiple 2D images, outdoor 3D semantic segmentation presents a more challenging setup in which only a single 2D image is paired with each pointcloud.

\subsection{Indoor 3D Semantic Segmentation}

In this section, first, we describe the datasets used to validate our TTT-KD, then explain our experimental setup, and lastly, we provide the results.

\subsubsection{Datasets.}

In our experiments, we use three different datasets of real indoor 3D scenes, \scannet~\cite{dai2017scannet}, \sdis~\cite{armeni2017joint}, and \matterportd~\cite{Matterport3D}.
These datasets are composed of several 3D scans of rooms from different buildings for which the reconstructed 3D pointcloud and a set of 2D images per pointcloud are available.
We follow the standard train, validation, and test splits of the datasets in our experiments.
For each point in the 3D scan, 3D coordinates, $[x,y,z]$, its normal, $[n_x,n_y,n_z]$, and color, $[r,g,b]$, are used as input to the models.

\subsubsection{Experimental setup.}

In this section, we describe the experimental setup used.
Additional details are provided in the appendix.

\paragraph{Tasks.}
We focus on two types of evaluations for our TTT-KD: \ac{ID} and \ac{ODD}.
For \ac{ID} evaluation, we train a model on the train split of a dataset and perform \ac{TTT} on the test split on the same dataset, while for \ac{ODD}, \ac{TTT} is performed on the test split of all other datasets in our evaluation setup. 
We report results by using the \ac{mIoU} evaluation metric for semantic segmentation.
For \ac{ODD} evaluations, since different datasets differ on the semantic labels used, we evaluate only the classes in which both the train and test dataset share.

\paragraph{Models.}
Our experiments use two different 3D backbones: a voxel-based and a point-based architecture.
As our voxel-based backbone, we choose the commonly used Minkowski34C~\cite{choy20194d}.
The point-based backbone is taken from Hermosilla~\etal~\cite{hermosilla2023pne} which is based on kernel point convolutions with Gaussian correlation functions.
As our foundation model, we use DINOv2~\cite{oquab2023dinov2}, in particular, the ViT-L/14 model with an embedding size of $1024$ features.

\paragraph{Testing.}
During training, we randomly rotate scenes along the up vector.
Therefore, during testing, we accumulate the logits over $8$ predictions of the same scene but rotated with different angles, covering $360$ degrees.
In the \ac{TTT} phase, we use \ac{SGD} without momentum and a large learning rate of $1$, and perform $100$ optimization steps for each rotated scene on the offline version of our algorithm, but only one for the online version.

\paragraph{Baselines.}
We train our models on the main segmentation task, \emph{\baseline}, and also using knowledge distillation as a secondary objective, \emph{\join}.
We compare their performance to the offline version of our algorithm, \emph{\ttkd}, and the online version, \emph{\ttkdo}.
Since, in the literature, there is no \ac{TTT} method proposed for the task of 3D semantic segmentation, we port several works from the image domain. 
Specifically, we compare our method with \emph{\tent}~\cite{wang2020tent}, \emph{\dua}~\cite{mirza2022dua}, and \emph{\tttcont}~\cite{chen2022contrastive}. 
While \tent uses entropy minimization at test-time, \dua updates the mean and standard deviation of the batch normalization layers.
\tttcont, on the other hand, leverages contrastive learning and momentum encoder to adapt the parameters of the model during testing.
Lastly, for \ac{ODD} experiments, as an upper bound, we provide the performance of an oracle model that has been trained on the same dataset as the test set, \emph{\oracle}.

\begin{table}[!t]
\caption{ 
Our method achieves large improvements not only on \ac{ODD} data but also on \ac{ID} setups, surpassing existing methods by a large margin.
Moreover, these results show that our algorithm is backbone agnostic, achieving comparable results for a point-based backbone, PNE~\cite{hermosilla2023pne}, and a voxel-based backbone, Mink~\cite{choy20194d}. }
\label{tbl:res_sem_seg}
\setlength{\tabcolsep}{3.5pt}
\begin{center}
\scriptsize
\begin{tabular}{llcccccc}
    \toprule
     & & \multicolumn{6}{c}{Test}\\
     \cmidrule(l{2pt}r{2pt}){3-8}
     \multicolumn{1}{l}{Train} & \multicolumn{1}{l}{Method} & \multicolumn{2}{c}{\scannet} & \multicolumn{2}{c}{\sdis} & \multicolumn{2}{c}{\matterportd}\\
     \cmidrule(l{2pt}r{2pt}){3-4} \cmidrule(l{2pt}r{2pt}){5-6} \cmidrule(l{2pt}r{2pt}){7-8}
     & & \multicolumn{1}{c}{PNE~\cite{hermosilla2023pne}} & \multicolumn{1}{c}{Mink~\cite{choy20194d}} & \multicolumn{1}{c}{PNE~\cite{hermosilla2023pne}} & \multicolumn{1}{c}{Mink~\cite{choy20194d}} & \multicolumn{1}{c}{PNE~\cite{hermosilla2023pne}} & \multicolumn{1}{c}{Mink~\cite{choy20194d}}\\

     \midrule

     \multirow{7}{*}{\rotatebox[origin=c]{90}{\scannet}}
     & \oracle & \cbgb \textcolor{OracleColor}{--} & \cbgb \textcolor{OracleColor}{--} & \cbgb \textcolor{OracleColor}{75.7} & \cbgb \textcolor{OracleColor}{77.4} & \cbgb \textcolor{OracleColor}{53.9} & \cbgb \textcolor{OracleColor}{52.3}\\
     
     & \baseline & \cbg 73.5 & \cbg 72.9 & 65.8 & 67.3 & \cbg 49.1 & \cbg 48.3 \\
     & \join & \cbg 74.3 \incr{0.8} & \cbg 73.9 \incr{1.0} & 66.5 \incr{0.7} & 71.5 \incr{4.2} & \cbg 50.4 \incr{1.3} & \cbg 48.7 \incr{0.4}\\
     
     & \tent & \cbg 71.1 \decr{2.4} & \cbg 68.8 \decr{4.1} & 46.8 \decr{19.0} & 70.5 \incr{3.2} & \cbg 47.1 \decr{2.0} & \cbg 44.1 \decr{4.2}\\
     & \dua & \cbg 73.9 \incr{0.4} & \cbg 73.1 \incr{0.2} & 64.0 \decr{1.5} & 70.6 \incr{3.3} & \cbg 48.9 \decr{0.2} & \cbg 46.9 \decr{1.4}\\
     
     & \tttcont & \cbg 73.8 \incr{0.3} & \cbg 72.4 \decr{0.5} & 67.2 \incr{1.4} & 72.3 \incr{5.1} & \cbg 50.2 \incr{1.1} & \cbg 48.4 \incr{0.1}\\
     
     & \ttkdo & \cbg 75.5 \incr{2.0} & \cbg 74.7 \incr{1.8} & \textbf{72.4 \incr{6.6}} & \textbf{73.7 \incr{6.4}} & \cbg 53.6 \incr{4.5} & \cbg 51.3 \incr{3.0}\\
     & \ttkd  & \cbg \textbf{76.4 \incr{2.9}} & \cbg \textbf{76.6 \incr{3.7}} & 70.4 \incr{4.6} & 73.1 \incr{5.8} & \cbg \textbf{56.6 \incr{7.5}} & \cbg \textbf{55.3 \incr{7.0}}\\
     
     \midrule

     \multirow{7}{*}{\rotatebox[origin=c]{90}{\sdis}}
     & \oracle & \cbgb \textcolor{OracleColor}{84.6} & \cbgb \textcolor{OracleColor}{84.2} & \cbgb \textcolor{OracleColor}{--} & \cbgb \textcolor{OracleColor}{--} & \cbgb \textcolor{OracleColor}{64.9} & \cbgb \textcolor{OracleColor}{66.0} \\
     
     & \baseline & 54.5 & 54.9 & \cbg 63.2 & \cbg 65.9 & 46.1 & 42.1 \\
     & \join &  55.5 \incr{1.0} & 56.1 \incr{1.2} & \cbg 65.4 \incr{2.2} & \cbg 66.8 \incr{0.9} & 47.0 \incr{0.9} & 42.8 \incr{0.7} \\
     
     & \tent & 56.0 \incr{1.5} & 54.6 \decr{0.3} & \cbg 53.0 \decr{10.2} & \cbg 66.1 \incr{0.2} & 45.6 \decr{0.5} & 43.4 \incr{1.3}\\
     & \dua & 59.0 \incr{4.5} & 57.6 \incr{2.7} & \cbg 67.3 \incr{4.1} & \cbg 65.5 \decr{0.4} & 46.7 \incr{0.6} & 44.1 \incr{2.0}\\
     
     & \tttcont & 58.0 \incr{3.5} & 57.5 \incr{2.6} & \cbg 65.4 \incr{2.2} & \cbg 65.6 \incr{0.3} & 46.7 \incr{0.6} & 46.4 \incr{4.3}\\
     
     & \ttkdo & 65.0 \incr{10.5} & 64.1 \incr{9.2} & \cbg 68.8 \incr{5.6} & \cbg 68.7 \incr{2.8} & 50.1 \incr{4.0} & 49.2 \incr{7.1}\\
     & \ttkd  & \textbf{69.9 \incr{14.4}} & \textbf{68.4 \incr{13.5}} & \cbg \textbf{71.7 \incr{8.5}} & \cbg \textbf{71.5 \incr{5.6}} & \textbf{53.2 \incr{7.1}} & \textbf{50.9 \incr{8.8}} \\
     
     \midrule

     \multirow{7}{*}{\rotatebox[origin=c]{90}{\mattershort}}
     & \oracle & \cbgb \textcolor{OracleColor}{73.5} & \cbgb \textcolor{OracleColor}{72.9} & \cbgb \textcolor{OracleColor}{77.9} & \cbgb \textcolor{OracleColor}{78.5} & \cbgb \textcolor{OracleColor}{--} & \cbgb \textcolor{OracleColor}{--} \\
     
     & \baseline & \cbg 49.0 & \cbg 45.4 & 59.2 & 58.6 & \cbg 55.2 & \cbg 53.8 \\
     & \join & \cbg 52.6 \incr{3.6} & \cbg 50.6 \incr{5.2} & 59.9 \incr{0.7} & 63.7 \incr{5.1} & \cbg 56.0 \incr{0.8} & \cbg 55.4 \incr{1.6} \\
     
     & \tent & \cbg 50.3 \incr{1.3} & \cbg 47.5 \incr{2.1} & 52.3 \decr{6.9} & 65.0 \incr{6.4} & \cbg 54.1 \decr{1.1} & \cbg 54.7 \incr{0.9}\\  
     & \dua & \cbg 52.7 \incr{3.7} & \cbg 50.7 \incr{5.3} & 64.5 \incr{5.3} & 63.8 \incr{5.2} & \cbg 56.0 \incr{0.8} & \cbg 54.7 \incr{0.9}\\
     
     & \tttcont & \cbg 55.2 \incr{6.2} & \cbg 53.4 \incr{8.0} & 60.8 \incr{1.6} & 69.7 \incr{11.1} & \cbg 56.0 \incr{0.8} & \cbg 54.3 \incr{0.5}\\
     
     & \ttkdo & \cbg 59.4 \incr{10.4} & \cbg 57.6 \incr{12.2} & 64.4 \incr{5.2} &  70.9 \incr{12.3} & \cbg 57.8 \incr{2.6} & \cbg 56.2 \incr{2.4}\\
     & \ttkd & \cbg \textbf{64.0 \incr{15.0}} & \cbg \textbf{62.6 \incr{17.2}} & \textbf{66.8 \incr{7.6}} & \textbf{75.4 \incr{16.8}} & \cbg \textbf{58.7 \incr{3.5}} & \cbg \textbf{56.5 \incr{2.7}}\\
     
    \bottomrule
\end{tabular}
\end{center}
\end{table}

\subsubsection{Results.}
The main experimental results and comparisons with all other methods are provided in \cref{tbl:res_sem_seg}.
In the following, we explain these results in detail. 

\paragraph{Joint-Training.}
First, we compare the performance of a \baseline model with our \join strategy.
Our results in \cref{tbl:res_sem_seg} show that for all datasets and both 3D backbones, joint training always provides an improvement over the \baseline model.
In some cases, this improvement is minor, such as in \scannet or \matterportd with an improvement of $0.8$ mIoU, but for other datasets the improvement is larger, as in \sdis or in \matterportd $\rightarrow$ \scannet with an improvement of $2.2$ and $3.6$ mIoU respectively.
We conjecture that the reason for the improvement is the \ac{KD} task acts as an additional regularizer.

\paragraph{In-distribution.}
When testing on \ac{ID} data, our algorithm provides significant improvements for all three datasets and all 3D backbones.
Our algorithm presents an improvement of $2.9$ and $3.7$ for \scannet, of $8.5$ and $5.6$ for \sdis, and $3.5$ and $2.7$ for \matterportd. 
Moreover, although smaller, the online version of our algorithm, \ttkdo, also presents significant gains on all datasets.

\paragraph{Out-of-distribution.}
When we look at the performance of the \baseline models when tested on \ac{ODD} data, as expected, the performance drops significantly when compared with an \oracle model trained on \ac{ID} data,
with large drops in performance as in \matterportd $\rightarrow$ \scannet with $24.5$ or in \sdis $\rightarrow$ \scannet with $30.1$.
Our \ttkd algorithm, on the other hand, is able to reduce this gap, increasing significantly the performance of all models and even obtaining better performance than the \oracle model as in the \scannet $\rightarrow$ \matterportd experiment.
Again, our online version, \ttkdo, also provides significant improvements but is smaller than our offline version.

\paragraph{Comparison to baselines.}
When compared to \tent, \dua, and \emph{\tttcont}, although these baselines can provide some adaptation, \ttkd has a clear advantage, surpassing them by a large margin.
We can see that \tent is not suited for the task of semantic segmentation, since it does not provide improvement in many of the configurations.
We hypothesize this is due to the mean and standard deviation of the batch normalization layers, which \tent computes independently for each test batch.
Since we are testing each scene independently, these estimates are not representative of the \ac{ODD} data, leading to a degradation of performance.
We can also see that \dua performs better than \tent, since it accumulates these parameters over several scenes, but still fails for some configurations.
Lastly, we can see that the more complex method \tttcont, performs better than both \dua and \tent but falls behind our \ttkd.
\begin{figure}[t]
    \centering
    \includegraphics[width=\linewidth]{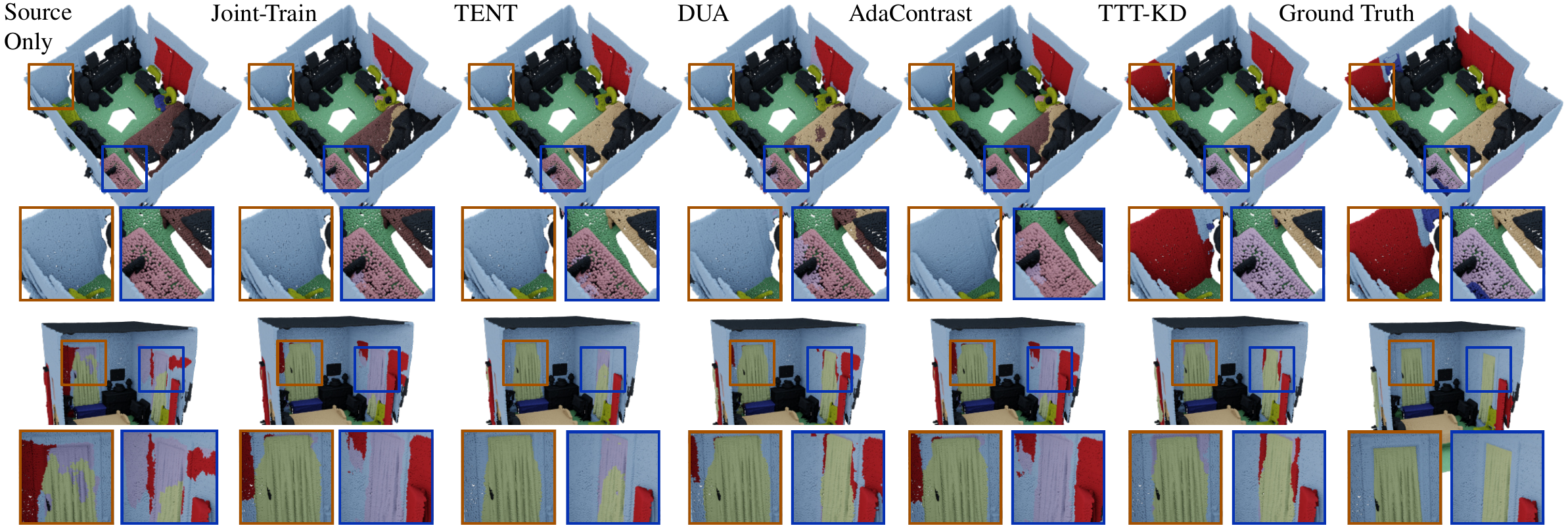}
    \caption{Qualitative results for two different \ac{ODD} tasks. The top row presents results for \matterportd $\to$ \scannet, while the bottom row presents results for \scannet $\to$ \matterportd.
    Although other methods are able to slightly adapt to the domain shifts, our \ttkd algorithm provides more accurate predictions.}
    \label{fig:examples}
\end{figure}
\cref{fig:examples} presents some qualitative results of these methods.

\paragraph{Backbone agnostic.}
When we analyze the performance of our method on different backbones, we see a consistent improvement in all setups.
This indicates that our method is independent of the 3D backbone used.

\subsection{Outdoor 3D Semantic Segmentation}

In this section, first, we briefly describe the datasets used, then the experimental setup, and lastly, we provide the results of the experiments.

\subsubsection{Datasets.}

In our experiments, we use two different autonomous driving datasets of real outdoor 3D scenes, \atwodtwo~\cite{a2d22020} and \skitti~\cite{behley2019iccv}. 
While the 3D pointclouds are obtained with LiDAR scans, the images are obtained from different cameras mounted on the vehicle.
Following Jaritz \etal~\cite{jaritz2019xmuda}, only the 2D images obtained from the front camera of the vehicle are used, and the 3D pointcloud is cropped by selecting only visible points from this camera.
For each point in the 3D pointcloud, only 3D coordinates are used.

\subsubsection{Experimental setup.}

For the task of outdoor 3D semantic segmentation, we use the same experimental setup as other \ac{UDA}  (\xmuda~\cite{jaritz2019xmuda}) and \ac{TTA} methods (\mmtta~\cite{shin2022mmtta}).
For additional details, we refer the reader to Jaritz \etal~\cite{jaritz2019xmuda}.

\paragraph{Tasks.}
We focus on a well-established and challenging task to measure the robustness of a model to \ac{ODD} data using \ac{mIoU} as our metric.
In this task, we train a model on the training set of the \atwodtwo dataset and perform \ac{TTT} on the test set of the \skitti.
Since the LiDAR scan used in the target domain is of higher resolution than the one used in the source domain, this task aims to measure the robustness of the 3D model to \ac{ODD} pointcloud data.

\paragraph{Model.}
As our 3D backbone, we use the same sparse convolution architecture as previous work~\cite{jaritz2019xmuda,shin2022mmtta}.
As our foundation model, we use again DINOv2~\cite{oquab2023dinov2}.

\paragraph{Testing.}
For testing, we use the same configuration as in the indoor 3D semantic segmentation tasks.
However, due to the large size of the dataset, we reduce the number of rotations to $4$, and the number of \ac{TTT} steps of our offline version to $25$. 
Moreover, we reduce the learning rate of the optimizer to $0.1$.

\paragraph{Baselines.}
As in the indoor datasets, we train several models: \emph{\baseline}, \emph{\join}, \emph{\ttkd}, \emph{\ttkdo}, and \emph{\oracle}.
Additional, we compare to the 3D backbone of the 2D-3D multi-modal \ac{TTA} method \emph{\mmtta}~\cite{shin2022mmtta}, to a \ac{TTA} version of \emph{\xmuda}~\cite{jaritz2019xmuda} as in Shin \etal~\cite{shin2022mmtta}, and, again, to  \emph{\tent}~\cite{wang2020tent}.

\begin{table}[t]
\caption{Results for the outdoor 3D semantic segmentation tasks. Our \ttkd algorithm significantly reduces the domain gap for \ac{ODD} compared to other methods.}
\label{tbl:res_sem_seg_out}
\setlength{\tabcolsep}{20pt}
\begin{center}
\scriptsize
\begin{tabular}{lc}
    \toprule
    \multirow{1}{*}{Method} & \multirow{1}{*}{\atwodtwo $\rightarrow$ \kitti}\\
     
    \midrule

    \oracle & \cbgb \textcolor{OracleColor}{73.8} \\

    \midrule
    
    \baseline & 35.8 \\
    \join & 41.6 \incr{5.8} \\

    \tent & 36.6 \incr{0.8} \\
    \xmuda & 38.0 \incr{2.2} \\
    \mmtta & 42.5 \incr{6.7} \\

    \midrule
    
    \ttkdo & \textbf{52.0 \incr{16.2}} \\
    \ttkd & 49.7 \incr{13.9} \\

    \bottomrule
\end{tabular}
\end{center}
\end{table}

\subsubsection{Results.}
The main results of this experiment are presented on \cref{tbl:res_sem_seg_out}.
In the following paragraphs, we analyze these results in detail.

\paragraph{Joint-Training.}
As in the indoor tasks, our \join strategy provides a significant improvement over the \baseline model.
Moreover, we can see that it is able to match and even surpass most of the baselines without performing any adaptation during testing, confirming that our \ac{KD} secondary task acts as regularizer, improving the generalization of the model. 

\paragraph{Out-of-distribution.}
When we analyze the performance of the \baseline models when tested on \ac{ODD}, we see again a significant performance drop when compared with an \oracle model trained on \ac{ID} data.
Our \ttkd algorithm, on the other hand, presents a large performance increase when compared to the \baseline.
The domain gap is reduced even more by our \ttkdo, achieving an increase of $45\,\%$ \ac{mIoU}.
We hypothesize that this is due to the reduced number of \ac{TTT} iterations of our offline version when compared to the number of iterations used for the indoor tasks. 

\paragraph{Comparison to baselines.}
When compared to commonly used baselines for \ac{TTA} in outdoor 3D semantic segmentation, we can see that our \ttkd and \ttkdo algorithms surpass them by a large margin.
While \mmtta is able to achieve reasonable good performance, \xmuda provides a small improvement over the \baseline model.
\tent, as in the indoor tasks, provides a marginal improvement.

\begin{figure}[t]
    \centering
    \includegraphics[width=\linewidth]{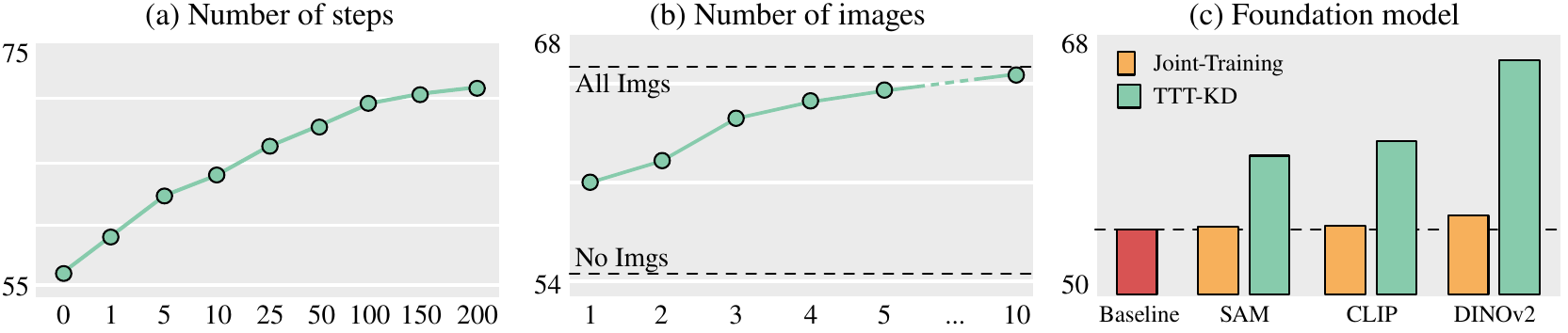}
    \caption{\textbf{Ablation studies. (a)} Improvement \wrt the number of \ac{TTT} step utilized per scene. \textbf{(b)} mIoU of the model \wrt to the number of images used in our \ttkd. \textbf{(c)} Comparison of DINOv2~\cite{oquab2023dinov2} to CLIP~\cite{radford2021learning} and SAM~\cite{kirillov2023segany} as our foundation model.}
    \label{fig:abl}
\end{figure}

\subsection{Ablation Studies}

In this section, we describe the ablation studies carried out to investigate the effect of different design choices on the performance of our algorithm.
Unless otherwise stated, due to computational reasons, all these ablations are performed on the task of indoor 3D semantic segmentation while adapting from \sdis $\rightarrow$ \scannet, with $25$ \ac{TTT}~steps using our offline setup.

\paragraph{Number of TTT steps.}
We measure the performance of our \ttkd algorithm in relationship to the number of \ac{TTT} steps and plot the results in \cref{fig:abl} (a).
We see that the \ac{mIoU} increases with the number of \ac{TTT} updates, and saturates at $200$ steps.
However, relative improvement is reduced after $100$ steps.

\paragraph{Number of images.}
Our algorithm relies on paired pointcloud and image data, which might be restrictive for some setups.
Therefore, we measure the performance of our method \wrt the number of images used for \ac{KD}.
\cref{fig:abl} (b) presents the results of this experiment.
We can see that even when only a single image is used, we can achieve a boost in \ac{mIoU} of $4.5$.
These results support the findings on the outdoor tasks, where also only one image is available for adaptation.
Moreover, we can see that the improvement saturates for $5$ images when the improvement obtained by including an additional image is reduced.

\paragraph{Foundation model.}
For a more generalizable \ac{TTT} approach, it must work with any of the readily available \textit{off-the-shelf} foundation models. 
To evaluate this, we experiment with different foundation models used for TTT-KD.
In this experiment, we compare the foundation model used in our main experiments, DINOv2~\cite{oquab2023dinov2}, to a CLIP~\cite{radford2021learning} model trained with 2\,B paired image-text data, and to a SAM~\cite{kirillov2023segany} model trained with 11\,M annotated images, and provide the results in \cref{fig:abl} (c).
Although previous works~\cite{oquab2023dinov2, darcet2023registers} have shown that DINOv2 provides better segmentation masks than CLIP, our \ttkd is also able to provide considerable performance gains while using CLIP in our pipeline.
Moreover, \ttkd is able to obtain similar results when the foundation model is trained with ground truth image segmentation masks, such as the SAM model.

\section{Limitations}
\label{sec:limitations}

Despite the number of benefits of our TTT-KD, it is not totally exempt from limitations.
Similar to other \ac{TTT} methods, the main limitation is the additional computation required for \ac{TTT} in comparison to simply evaluating a network with frozen weights.
Our \textit{standard} \ttkd requires processing each image with a foundation model and then performing several optimization steps for each scene.
Still, this could be drastically reduced with our online version, which only optimizes the network weights for a single step, as other \ac{TTA} methods, and shows considerable improvements even surpassing the offline version for some datasets. 
Another limitation is that our algorithm relies on the robustness of the foundation model to \ac{ODD} data.
From the experiments presented in the paper, and additional experiments provided in the appendix, we concluded that DINOv2~\cite{oquab2023dinov2} is robust to domain shifts.
However, we acknowledge that there might be some cases where the foundation model used does not present such robustness.
In such cases, existing \ac{TTT} or \ac{TTA} approaches for 2D images could be used to adapt the foundation model to this new domain.
\section{Conclusions}
\label{sec:conclusions}

Our TTT-KD is the first test-time training method proposed for the task of 3D semantic segmentation, which proposes to use knowledge distillation from foundation models as a self-supervised auxiliary objective to adapt the network weights individually for each test sample as it is encountered.
Our experiments show that TTT-KD can be used with any off-the-shelf foundation model and multiple different 3D backbones.
Furthermore, our method provides impressive performance gains while adapting to both in-distribution and out-of-distribution test samples when evaluated on multiple different benchmarks.

\clearpage
{
    \bibliographystyle{splncs04}
    \bibliography{main}
}

\clearpage
\appendix

\section{Additional Experiments}

\subsection{\scannet Benchmark}

\begin{wraptable}{r}{0.5\textwidth}
\vspace{-1.1cm}
\setlength{\tabcolsep}{6pt}
\caption{Results on the test and validation sets of \scannet. \emph{Italic} indicates a pre-training strategy or additional data was used.}
\label{tbl:scannet-res}
\begin{center}
\scriptsize
\begin{tabular}{lccrr}
    \toprule
    Method & Res. & Val. & Test\\
    \midrule
    MinkowskiNet~\cite{choy20194d} & 2cm& 72.2 & 73.6 \\
    Point Transf. V2~\cite{wu2022pointtrans} & 2cm & {75.4} & {75.2} \\
    PNE~\cite{hermosilla2023pne} & 2cm & 74.9 & {75.5} \\
    OctFormer~\cite{Wang2023octformer} & 1cm & {75.7} & {76.6} \\
    Point Transf. V3~\cite{wu2024ptv3} & 2cm & 77.5 & {77.9} \\
    \hspace{2.5 mm}+ \emph{PointPrompt}~\cite{wu2024ppt} & 2cm& \textbf{78.6} & \textbf{79.4} \\
    \emph{PointPrompt}~\cite{wu2024ppt} & 2cm & 76.4 & 76.6 \\
    \emph{Swin3D}~\cite{yang2023swin3d} & 2cm & {77.5} & {77.9} \\
    \emph{PonderV2}~\cite{zhu2023ponderv2} & 2cm & 77.0 & \underline{78.5} \\
    \midrule
    \join & \multirow{2}{*}{2cm} & 75.7 & -- \\
    \ttkd & & \underline{77.6} & 77.3 \\
    \bottomrule
\end{tabular}
\end{center}
\vspace{-.5cm}
\end{wraptable}
Since our \ttkd provides a clear improvement on \ac{ID} data, we compare our strategy to SOTA models on the \scannet benchmark for 3D semantic segmentation.
For this experiment, we decrease the resolution of the subsampling step to $2\,$cm and keep the same data augmentation techniques as in our main experiments.
\cref{tbl:scannet-res} presents the result of this experiment.
We can see that our model trained jointly with the self-supervised task provides a competitive performance, obtaining competitive validation accuracy over the supervised methods, only surpassed by the recent Point transformer v3~\cite{wu2024ptv3}, but falling behind methods using pre-training strategies or additional data.
However, when we perform \ac{TTT} with our \ttkd algorithm, we can see that it outperforms most of the existing methods on the validation set, only surpassed by the recent Point transformer v3~\cite{wu2024ptv3} trained with additional data, while achieving competitive performance on the test set.
Please note that historically TTT methods have been proposed to adapt to OOD data only~\cite{mirza2023mate, sun2020test, gandelsman2022test}, however, our \ttkd also shows impressive performance gains while adapting to \ac{ID} data. 

\begin{figure}
    \centering
    \includegraphics[width=\linewidth]{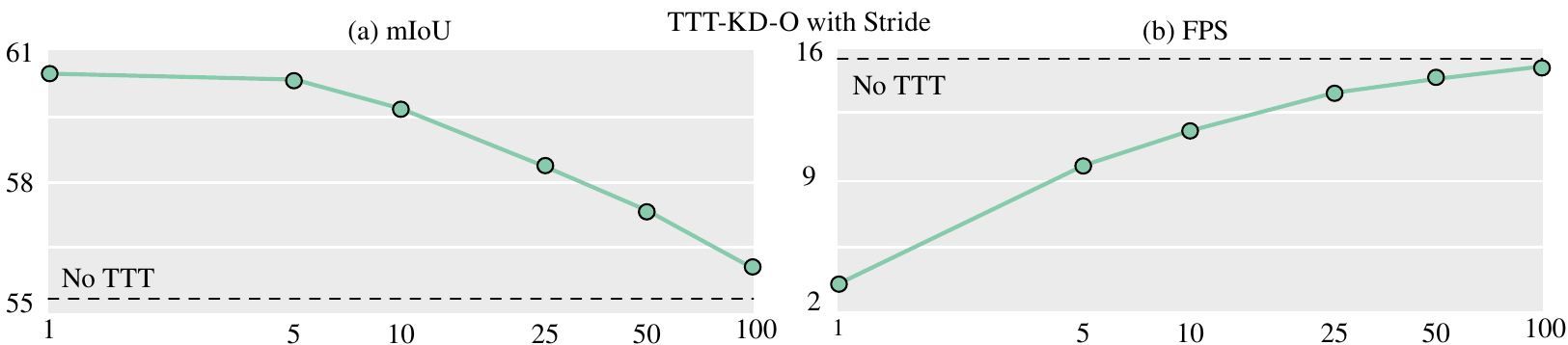}
    \caption{Results obtained when updating the model's parameters using our \ttkdo algorithm with stride $s$, \ie only performing a {TTT} step every $s$ predictions.
    Results show that even for large stride values our algorithm produces significant improvements while maintaining low computational cost.}
    \label{fig:stride}
\end{figure}

\subsection{Computational Cost}
TTT-KD is computationally very inexpensive. 
All our experiments are performed on a single NVIDIA $A6000$ GPU, requiring $2$\,Gb memory and, on average, $177$\,ms for each TTT step in the \scannet dataset, plus $210$\,ms for each image processed by the foundation model, DINOv2 ViT-L.
However, if speed is a concern for some applications, we can reduce the frequency of parameter updates in our online {TTT} setup.
\cref{fig:stride} (a) shows, for the \sdis $\to$ \scannet setup with a single image per scene, the {mIoU} for a different number of predictions in between each parameter update, \ie a stride of $5$ indicates that the parameters of the model are updated every $5$ scene predictions.
We can see that even if our \ttkdo is sparsely applied, this still leads to a significant improvement.
Moreover, \cref{fig:stride} (b) shows the predictions per second of our method for different values of stride.
We can see that, even when we apply our TTT step every $5$ predictions, we can perform almost $10$ predictions per second. 

Additionally, since the main cost of our algorithm is the image processing with the foundation model, we could use a smaller model, such as the distilled ViT-S version of DINOv2, reducing the time from $210$\,ms per image to $25$\,ms.

\subsection{Robustness of Foundation Model}
One of the underlying goals of our \ttkd is to exploit the world model inherently learned by foundation models during their large-scale pre-training and avoid fine-tuning to save computation overhead during TTT. 
Furthermore, DINOv2 reports strong robustness to distribution shifts,~\eg on ImageNet-C (IN-C) benchmark~\cite{hendrycks2019robustness}. 
To verify this in our setup, we perturb the images with the distribution shifts proposed in the IN-C benchmark and provide the final adaptation results while adapting from S3DIS $\to$ SCANNET in \cref{tab:abl-corrupt}.
We observe no performance drop by adding these corruptions.
This might be because we down-sample the images by $4 \times$ before feeding them to DINOv2, which alleviates the effect of distortions provoked by these perturbations.

\begin{table}
    \setlength{\tabcolsep}{4pt}
    \centering
    \caption{TTT-KD results with distribution shifted images.}
    \label{tab:abl-corrupt}
    \begin{tabular}{cccccc}
       \toprule
       & & \multicolumn{4}{c}{ImageNet-C Corruptions}\\
       \cmidrule{3-6}
       \textit{Source-Only} & Clean & Shot & Blur & Bright & Jpeg \\
       \hline
       55.5 & 66.6 & 66.6 & 66.7 & 66.6 & 66.4 \\
       \hline
    \end{tabular}
\end{table}

\subsection{Additional Qualitative Results}
\begin{figure}[t]
    \centering
    \includegraphics[width=\linewidth]{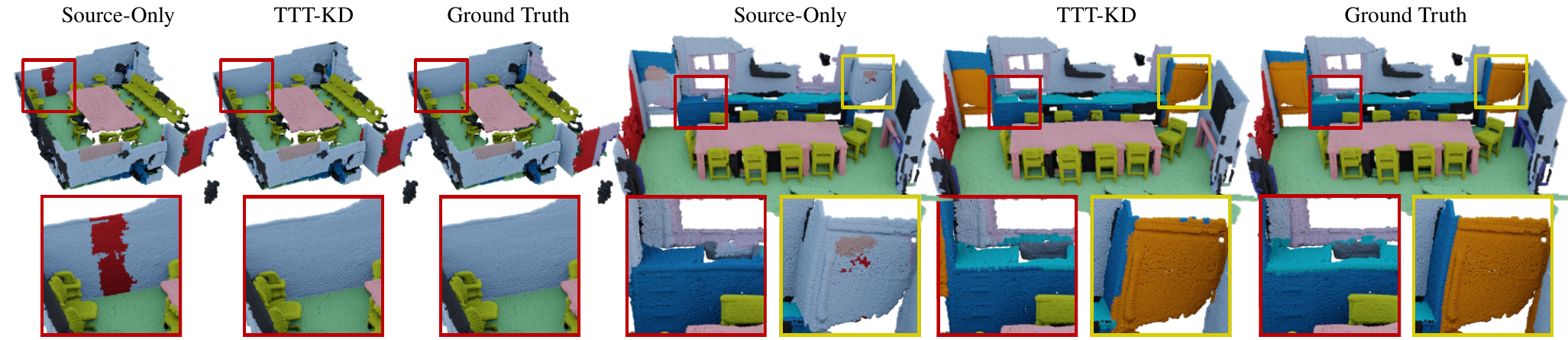}
    \caption{Additional qualitative results of the Mink~\cite{choy20194d} network trained and tested on the \scannet dataset.
    We can see that even if training and testing data are from the same distribution, our \ttkd algorithm also improves the predictions of the model.}
    \label{fig:add-qual}
\end{figure}

Fig.~\ref{fig:add-qual} provides additional qualitative results.

\section{Datasets}

\subsection{Indoor 3D Semantic Segmentation}

\paragraph{\scannet~\cite{dai2017scannet}.}
This dataset is composed of $1,513$ real 3D scans of different rooms, for which each point in the scan is classified among $20$ different classes.
Moreover, for each scan, a set of images used for the 3D reconstruction is provided.
The dataset is divided into two splits, $1,201$ rooms for training and $312$ rooms for validation.
Moreover, data samples from an additional $100$ rooms are provided as a test set for benchmarking where ground truth labels are not available.
In our experiments, we train the models on the train set and report performance on the validation set.

\paragraph{\sdis~\cite{armeni2017joint}.}
This dataset provides dense pointclouds of $271$ different rooms from $6$ large-scale areas.
Each point in the dataset is classified among $13$ different classes.
The dataset also provides the 2D images used for the 3D reconstruction.
Following previous works~\cite{thomas2019KPConv,hengshuang2021pointtrans}, we use data collected in areas $1$, $2$, $3$, $4$, and $6$ for training and data collected in area $5$ for testing.

\paragraph{\matterportd~\cite{Matterport3D}.}
Matterport3D provides pointclouds from $90$ building-scale scenes, each composed of multiple regions or areas, and a set of images used for the 3D reconstruction.
Each point within the dataset is classified among $21$ classes.
We follow the official splits and use $61$ scenes for training, $11$ for validation, and $18$ for testing.
We report the performance of the models on the test split.

\paragraph{Data augmentations.}
Although it is common practice to use specific data augmentations for each data set, this might lead to a bigger generalization gap when evaluating on \ac{ODD} datasets.
Therefore, for a fair baseline, we use a fixed set of data augmentations for all the data sets.
We use random mirror of the $X, Y$ axes, random rotations around the up vector, random scaling, elastic distortion, jitter of point coordinates, random crop, random translation, random adjustments of brightness and contrast of the point's colors, and RGB shift. 
Moreover, we subsample the scene using a voxel size of $4\,$cm.
Lastly, we mix two scenes~\cite{Nekrasov213DV} with a probability equal to $0.5$.

\paragraph{Semantic labels.}
Tbl.~\ref{tbl:labels} presents the list of classes per each dataset.
We can see that \scannet and \matterportd share almost all classes with the exception of \emph{ceilling}.
However, with \sdis the number of shared classes is only $8$ with \scannet and $9$ with \matterportd.

\begin{table*}
\caption{List of all semantic labels used on our experiments and their presence in each individual dataset.}
\label{tbl:labels}
\setlength{\tabcolsep}{2.2pt}
\begin{center}
\footnotesize
\begin{tabular}{lccccccccccccccccccccccccc}
    \toprule
     & \rotatebox[origin=l]{90}{Bathtub}
     & \rotatebox[origin=l]{90}{Beam}
     & \rotatebox[origin=l]{90}{Bed}
     & \rotatebox[origin=l]{90}{Board}
     & \rotatebox[origin=l]{90}{Bookshelf}
     & \rotatebox[origin=l]{90}{Cabinet}
     & \rotatebox[origin=l]{90}{Ceiling}
     & \rotatebox[origin=l]{90}{Chair}
     & \rotatebox[origin=l]{90}{Clutter}
     & \rotatebox[origin=l]{90}{Column}
     & \rotatebox[origin=l]{90}{Counter}
     & \rotatebox[origin=l]{90}{Curtain}
     & \rotatebox[origin=l]{90}{Desk}
     & \rotatebox[origin=l]{90}{Door}
     & \rotatebox[origin=l]{90}{Floor}
     & \rotatebox[origin=l]{90}{Otherfurniture}
     & \rotatebox[origin=l]{90}{Picture}
     & \rotatebox[origin=l]{90}{Refrigerator}
     & \rotatebox[origin=l]{90}{Shower curtain}
     & \rotatebox[origin=l]{90}{Sink}
     & \rotatebox[origin=l]{90}{Sofa}
     & \rotatebox[origin=l]{90}{Table}
     & \rotatebox[origin=l]{90}{Toilet}
     & \rotatebox[origin=l]{90}{Wall}
     & \rotatebox[origin=l]{90}{Window}
     \\
    \midrule
    \sdis & & \checkmark & & \checkmark & \checkmark & & \checkmark & \checkmark & \checkmark & \checkmark & & & & \checkmark & \checkmark & & & & & & \checkmark & \checkmark & & \checkmark & \checkmark\\
    \scannet & \checkmark & & \checkmark & & \checkmark & \checkmark & & \checkmark & & & \checkmark & \checkmark & \checkmark & \checkmark & \checkmark & \checkmark & \checkmark & \checkmark & \checkmark & \checkmark & \checkmark & \checkmark & \checkmark & \checkmark & \checkmark\\
    \mattershort & \checkmark & & \checkmark & & \checkmark & \checkmark & \checkmark & \checkmark & & & \checkmark & \checkmark & \checkmark & \checkmark & \checkmark & \checkmark & \checkmark & \checkmark & \checkmark & \checkmark & \checkmark & \checkmark & \checkmark & \checkmark & \checkmark\\
    \bottomrule
\end{tabular}
\end{center}
\end{table*}

\subsection{Outdoor 3D Semantic Segmentation}

\paragraph{\atwodtwo~\cite{a2d22020}.} 
This dataset contains multiple outdoor driving scenes of paired point clouds and images.
The point clouds are obtained from three overlapping low-resolution LiDAR scans with $16$ layers each that generate a rather sparse point cloud. 
Following the experimental setup of xMUDA~\cite{jaritz2019xmuda}, we only use the points that are projected on the front camera of the vehicle.
For training, we use the standard split used in xMUDA, resulting in more than $27$\,K scenes.

\paragraph{\skitti~\cite{behley2019iccv}.}
Similar to \atwodtwo, this dataset also contains multiple outdoor driving scenes of paired point clouds and images.
The point cloud in this dataset were obtained with a single high-resolution LiDAR scan with $64$ layers.
Again, we follow the experimental setup of xMUDA~\cite{jaritz2019xmuda}, and only use the points that are projected on the front camera of the vehicle.
For testing on this dataset, we use the split used in xMUDA, resulting in more than $4$\,K scenes.

\paragraph{Semantic labels.}
As in xMUDA~\cite{jaritz2019xmuda}, we select $10$ shared semantic labels between the two datasets, \atwodtwo and \skitti:
car, truck, bike, person, road, parking, sidewalk, building, nature, and other-objects.
For more details on class mapping, we refer the reader to xMUDA~\cite{jaritz2019xmuda}.

\section{Training Details}

\subsection{Indoor 3D Semantic Segmentation}

\paragraph{Training.}
We use AdamW~\cite{loshchilov2018decoupled} as our optimizer with a maximum learning rate of $0.005$ and a OneCycleLR learning rate scheduler~\cite{smith2017fastconv} with an initial division factor of $10$, and a final factor of $1000$.
To prevent overfitting, we use a weight decay value of $0.0001$ and label smoothing for the segmentation task with a value of $0.2$.

\paragraph{Test-Time Training.}
For {TTT}, we use {SGD} without momentum and a learning rate equal to $1.0$.
We optimize $100$ steps for each scene before performing the final prediction.

\paragraph{Number of images.}
For \scannet, we used the images provided in the $25$\,K subset, where images from the RGB-D video sequence are taken at intervals of $100$, generating $16$ images per scene on average.
For \matterportd and \sdis we select the available images for each scene from the provided 2D data.
If more than $50$ images for each scene are available, we select randomly $50$.

\paragraph{Number of samples for KD loss.}
Since the knowledge distillation loss requires large amounts of GPU memory due to the large feature dimensions of the embeddings, we randomly sample $2048$ points per scene during training.
During testing, since we work with only one scene at a time, we increase this number of samples to $16,384$ samples.

\subsection{Outdoor 3D Semantic Segmentation}
For the outdoor 3D semantic segmentation task, we use the same training as setup as xMUDA~\cite{jaritz2019xmuda}.
However, due to the large feature dimensions of the image foundation models, for the knowledge distillation loss, we use the same number of samples as in the indoor 3D semantic segmentation tasks.

\section{TTT-KD Algorithm}
In Alg.~\ref{alg:tttkd} we present the detailed \ttkd algorithm, consisting of the joint-training, the {TTT}, and inference phases.

\begin{algorithm}[t]
    \caption{\ttkd algorithm.}\label{alg:tttkd}
    \textbf{Input: }
    Training data $\mathcal{S} = \{ (\mathcal{X}, \mathcal{F}, \mathcal{Y}, \mathcal{I}, \mathcal{U}) \}$ \\
    \Indp \Indp \hspace{.2 mm} Test sample $(\mathcal{X}^{\prime}, \mathcal{F}^{\prime}, \mathcal{I}^{\prime}, \mathcal{U}^{\prime})$\\
    \hspace{.2 mm} 3D Backbone $(\psi_{3D})$\\
    \hspace{.2 mm} 2D Foundation Model $(\phi_{2D})$\\
    \hspace{.2 mm} Projectors $(\rho_{\mathcal{Y}}, \rho_{2D})$\\
    \Indm \Indm \textbf{Result: }Prediction $(\mathcal{Y}^{\prime})$
    
    \Begin{
        {\color{AlgCommentColor}\# Joint-Training}\\
        \For{numEpochs}{
            $(\mathcal{X}, \mathcal{F}, \mathcal{Y}, \mathcal{I}, \mathcal{U}) = $ sampleBatch$(\mathcal{S})$\\
            $F^{3D} = \psi_{3D}(\mathcal{X}, \mathcal{F})$\\
            $\hat{\mathcal{Y}} = \rho_{\mathcal{Y}}(F^{3D})$\\
            $\hat{F}^{2D} = \rho_{2D}(F^{3D})$\\
            $F^{2D} = \phi_{2D}(\mathcal{I})$\\
            $\mathcal{L} = \mathcal{L}_{\mathcal{Y}}(\hat{\mathcal{Y}}, \mathcal{Y}) + \mathcal{L}_{2D}(\hat{F}^{2D}, F^{2D}, \mathcal{U})$\\
            $\mathcal{L}$.backward()\\
            optimizer.step()\\
        }

        $\mathcal{Y}^{\prime} = 0$\\
        \For{numRotations}{
            $\hat{\mathcal{X}}^{\prime} = $ Rotate$(\mathcal{X}^{\prime})$\\
            {\color{AlgCommentColor}\# Test-Time Training}\\
            \For{numTTTSteps}{
                $\hat{F}^{2D} = \rho_{2D}(\psi_{3D}(\hat{\mathcal{X}}^{\prime}, \mathcal{F}^{\prime}))$\\
                $F^{2D} = \phi_{2D}(\mathcal{I}^{\prime})$\\
                $\mathcal{L} = \mathcal{L}_{2D}(\hat{F}^{2D}, F^{2D}, \mathcal{U}^{\prime})$\\
                $\mathcal{L}$.backward()\\
                optimizer.step()\\
            }
            {\color{AlgCommentColor}\# Inference}\\
            $\mathcal{Y}^{\prime} \mathrel{+}= \rho_{\mathcal{Y}}(\psi_{3D}(\mathcal{X}^{\prime}, \mathcal{F}^{\prime}))$
        }
    }
    
\end{algorithm}

\end{document}